\documentclass{bmvc2k}

\usepackage{booktabs}
\usepackage{multirow}
\usepackage{array}
\usepackage{hhline}
\usepackage{subcaption}
\usepackage{floatrow}
\usepackage{bbm}

\newcommand{\cam}[1]{#1}

\title{View-consistent 4D Light Field\\Depth Estimation}

\addauthor{Numair Khan}{numair_khan@brown.edu}{1}
\addauthor{Min H. Kim}{minhkim@vclab.kaist.ac.kr}{2}
\addauthor{James Tompkin}{james_tompkin@brown.edu}{1}

\addinstitution{
 Brown University, 
 USA
}
\addinstitution{
 KAIST,
 South Korea
}

\runninghead{Khan, Kim, Tompkin}{View-Consistent 4D Light Field Depth Estimation}

\begin{document}

\maketitle

\vspace{-0.25cm}
\begin{abstract}
\cam{We propose a method to compute depth maps for every sub-aperture image in a light field in a view consistent way.}
Previous light field depth estimation methods typically estimate a depth map only for the central \cam{sub-aperture} view, and struggle with view consistent estimation. 
\cam{Our method} precisely defines depth edges via EPIs, then we diffuse these edges spatially within the central view. These depth estimates are then propagated to all other views in an occlusion-aware way. Finally, disoccluded regions are completed by diffusion in EPI space. 
\cam{Our method} runs efficiently with respect to both other classical and \cam{deep learning-based} approaches, and achieves competitive quantitative metrics and qualitative performance on both synthetic and real-world light fields.
\vspace{-0.2cm}
\end{abstract}


\section{Introduction}
\label{sec:intro}
\vspace{-0.15cm}

Light fields allows high-quality depth estimation of fine detail by aggregating disparity information across many \cam{sub-aperture} views.
This typically results in a depth map for the central view of a light field only. However, in principle we can estimate depth for every pixel in the light field. Some applications require this ability, such as when editing a light field photograph when every output view will be seen on a light field display. Estimating depth reliably for disoccluded regions is difficult because it requires aggregating information from fewer samples. Few existing methods estimate depth for every light field pixel. These are typically computationally expensive~\cite{wanner2012, zhang2016} or not strictly occlusion aware. Jiang et al.~\cite{jiang2019,jiang2018} presented the first practical view consistent method based on deep learning. 

We present a counterpart first principles method with no learned priors, which produces comparable or better accuracy and view consistency than the current state of the art while being 2--4$\times$ faster.
%
%
Our method is based around estimating accurate view-consistent disparity at edges, and then completing an occlusion-aware diffusion process to fill in missing regions. 

\cam{Depth diffusion is a long-standing problem in which it is difficult to ensure both consistency and correctness in disoccluded regions. Our key contribution is an angular inpainting method that ensures depth consistency by design, while accounting for the visibility of points in disoccluded regions. In this way, we avoid the problem of trying to constrain or regularize view consistency after estimating depth spatially, and so can maintain efficiency.}


Our code will be released as open source software at \href{https://visual.cs.brown.edu/lightfielddepth}{visual.cs.brown.edu/lightfielddepth}.
\section{Related Work}
\label{sec:relatedwork}
\vspace{-1mm}

The regular structure of an Epipolar Plane Image (EPI) obviates the need for extensive angular regularization and, thus, many light field operations seek to exploit it. Khan et al.'s~\cite{khan2019} light field superpixel algorithm operates in EPI space to ensure a view consistent segmentation. The depth information implicit within an EPI is useful for disparity estimation algorithms. Zhang et al.~\cite{zhang2016} propose an EPI spinning parallelogram operator for this purpose. This operator is similar in respects to the large Prewitt filters of Khan et al.~\cite{khan2019} but has a larger support, and provides more accurate estimates. A related method is presented by To\v{s}i\'{c} and Berkner~\cite{tosic2014} who create light field scale-depth spaces through convolution with a set of specially adapted kernels. Wang et al.~\cite{wang15,wang16} exploit the angular view presented by an EPI to address the problem of occlusion. Tao et al.'s \cite{tao2013} work uses both correspondence and defocus in a higher-dimensional EPI space for depth estimation. 

Beyond EPIs, Jeon et al.'s~\cite{jeon2015} method exploits the relation between defocus and depth too. They shift light field images by small amounts to build a subpixel cost volume. Chuchwara et al.~\cite{chuchvara2020} present an efficient light-field depth estimation method based on superpixels and PatchMatch~\cite{barnes09} that works well for wide-baseline views. Efficient computation is also addressed by the work of Holynski and Kopf~\cite{holynski2018}. Their method estimates disparity maps for augmented reality applications in real-time by densifying a sparse set of point depths obtained using a SLAM algorithm. Chen et al.~\cite{chen2018} estimate  occlusion boundaries  with superpixels in the central view of a light field to regularize the depth estimation process.

With deep learning, methods have sought to bypass the large number of images in a light field by learning `priors' that guide the depth estimation process. Huang et al.'s~\cite{huang18} work can handle an arbitrary number of uncalibrated views. Alperovich et al.~\cite{alperovich2018} showed that an encoder-decoder architecture can be used to perform light field operations like intrinsic decomposition and depth estimation for the central cross-hair of views. As one of the few depth estimation methods that operates on every pixel in a light field, Jiang et al.~\cite{jiang2018,jiang2019} generate disparity maps for the entire light field and enforce view consistency. However, as their method uses low-rank inpainting to complete disoccluded regions, it fails to account for occluding surfaces in reprojected depth maps. Our method uses occlusion-aware edges to guide the inpainting process and so captures occluding surfaces in off-center views. 
\section{Occlusion-aware Depth Diffusion}
\label{sec:method}
\vspace{-1mm}

A naive solution to estimate per-view disparity might be to attempt to compute a disparity map for each \cam{sub-aperture} view separately. However, this is typically challenging for edge views and is highly inefficient, not only in terms of redundant computation but also due to the spatial domain constraints or regularization that must be added to ensure that depth maps are mutually consistent across views.
Another simple approach might be to calculate a disparity map for a single view, and then reproject it into all other views. However, this approach fails to handle scene points that are not visible in the single source view. Such points cause holes in the case of disocclusions, or lead to inaccurate disparity estimates when the points lie on an occluding surface. While most methods try to deal with the former case through inpainting, for instance via diffusion, the latter scenario is more difficult to deal with as the occluding surface may have a depth label not seen in the original view. Thus, techniques like diffusion are insufficient on their own without additional guidance. 

Our proposed method deals with this issue of depth consistency in subviews of light fields via an occlusion-aware diffusion process. We estimate sparse depth labels at edges that are explicitly defined across views, and then efficiently determine their visibility in each \cam{sub-aperture} view. Given occlusion-aware edges which persist across views, these edge depth labels can be used as more reliable guides for filling any holes in reprojected views. Since the edge depth labels are not restricted to the source view, we capture any occluding surfaces not visible in the source view. This avoids the aforementioned problem of unseen depth labels. In addition, by performing our inpainting step in the angular rather than the spatial domain of the light field, we improve cross view consistency and occlusion awareness.

\paragraph{Edge Depth \& Visibility Estimation}
\label{sec:method-epi-depth}
To begin, we estimate depth labels at all edges in the light field using the EPI edge detection algorithm proposed by Khan et al.~\cite{khan2019}.
This filters each EPI with a set of large Prewitt filters, and then estimates a representation of each EPI edge as a parametric line. Lines in EPI space correspond to scene points, with the slope of the line being proportional to the depth of the point. Therefore, this line representation captures both position and disparity information of scene edges.

However, the parametric definition implies the line exists in all views, and, hence, the representation does not contain any visibility information for the point across light field views. We can address this by looking at local structure around the edge for point samples along the line: the point is visible if the local gradient matches the global line direction. We define a gradient-alignment-based visibility score for each sample, then threshold this score to decide which part of the line is actually visible in a given view~\cite{khan2020}.

Given an EPI line~$l$, we sample it at $n$ locations to obtain a set of point samples $S_l = \{x_i, y_i\}$ on the EPI $I$. Each sample $s_i$ corresponds to a projection of the original point in light field view $i$, whose visibility can be determined as:
%
%
\vspace{-0.5mm}
\begin{align}
    v_l(i) = \mathbbm{1}\left( \frac{\nabla I(s_i) (\nabla l)^T}{\lVert \nabla I(s_i) \rVert \lVert \nabla l \rVert} > cos(\tau_v) \right),
\end{align}
\vspace{-0.5mm}%
where $\mathbbm{1}$ denotes the characteristic function of a set beyond the visibility threshold, $\nabla I$ is the image gradient, $\nabla l$ is the direction perpendicular to the line $l$, and $\tau = \pi/13$.
%

\cam{The characteristic function was proposed by~\cite{khan2020} to generate a central view disparity map, and so only central view points were kept.} 
\cam{By extending this idea to the entire light field,}
we retain all points along with their depth and visibility information, and use them to guide the disparity propagation into disoccluded and occluded regions. Since disparity directly at edges can often be ambiguous due to image resolution limits, the points are offset along the image gradient so that they lie on an actual surface. This is completed using the two-way propagation scheme of~\cite{khan2020}, which also allows a disparity map for the central view to be generated using dense diffusion of the sparse point depths.

\paragraph{Cross-hair View Projection}
The EPI line-fitting algorithm works on EPIs in the central cross-hair views---that is, the central row and column of light field images. While it is possible to run it on other rows and columns, this can become expensive, and the central set is usually sufficient to detect visible surfaces in the light field~\cite{wanner2013}. Hence, we project the estimated disparity map from the center view into all views along the cross-hair. Since gradients at depth edges in the estimated disparity map are not completely sharp, this leads to some edges being projected onto multiple pixels in the target view. We deal with this by sharpening the edges of the disparity map before projection, as shown in Shih et al.~\cite{Shih2020}, using a weighted median filter~\cite{ma2013} with parameters $r = 7$ and $\epsilon = 10^{-6}$. \cam{Omitting this step can cause inaccurate estimates around strong depth edges. The result is not very sensitive to parameters $r$ and $\epsilon$ since most parameter settings will target the error-prone strong edges.}

\begin{figure}[t]
\centering
{\includegraphics[width=1.0\linewidth]{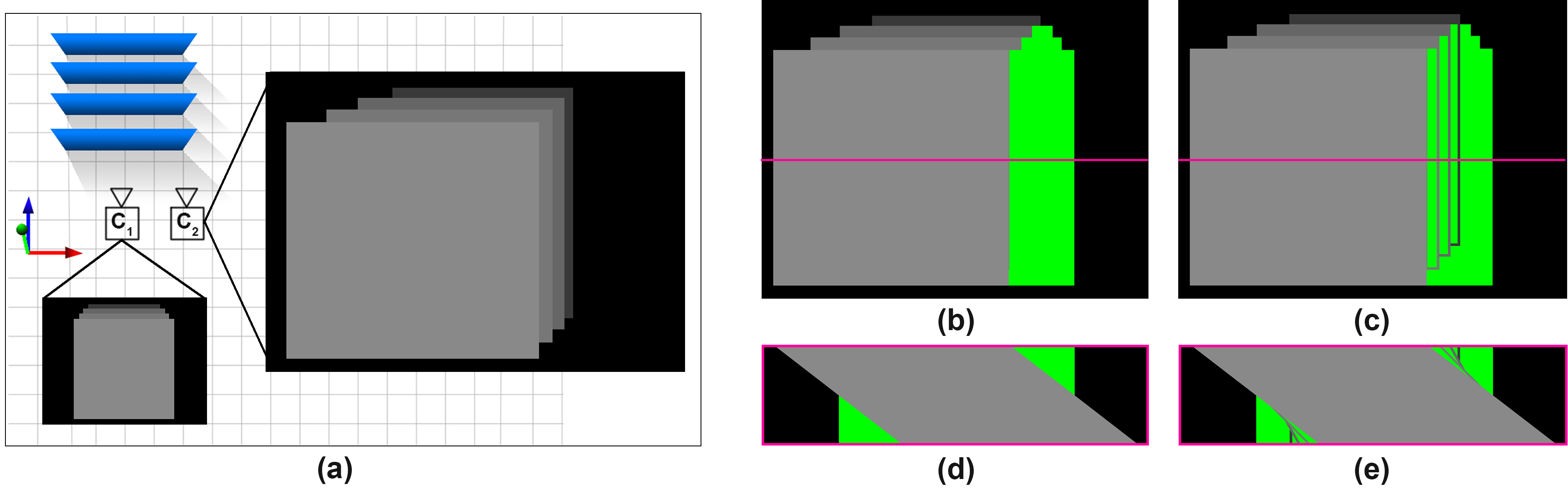}}
\vspace{-0.75cm}
{\caption{\textbf{(a)} Ground-truth disparity maps for a scene from two different camera positions $C_1$ and $C_2$. \textbf{(b)} Naively attempting to generate the output of $C_2$ by reprojecting $C_1$ results in large holes, shown here in green. \textbf{(c)} Our method uses depth edges to guide disparity propagation in such disoccluded regions. The EPIs corresponding to the highlighted row are shown in \textbf{(d)} and \textbf{(e)}. The EPI in \textbf{(e)} \cam{constitutes} our depth EPI $D_o$.}
\label{fig:epi-diffusion}}
\end{figure}

\paragraph{Angular Inpainting}
After depth reprojection, we must deal with the two problems highlighted in the overview: inpainting holes, and accounting for occluding surfaces in off-center \cam{sub-aperture} views. We tackle this by using the edges from Section \ref{sec:method-epi-depth} to guide a dense diffusion process. Moreover, we ensure view consistency by performing diffusion in EPI space.

\cam{The EPI lines from the first stage constitute a set $L$ of cross-view edge features. $L$ is robust to occlusions in a single view as it exists in EPI space. As such, $L$ provides occlusion-aware sparse depth labels to guide dense diffusion in EPI space. Diffusion in EPI space has the added advantage of ensuring view consistency.}

Let $D_o$ represent an angular slice of the disparity maps with values reprojected from the center view and with propagation guides (Figure~\ref{fig:epi-diffusion}). Then, we formulate diffusion as a constrained quadratic optimization problem:
\begin{align}
  \hat D = \underset{D}{\mathrm{argmin}} \sum_{p \in D} E_d(p) + \sum_{(p, q) \in \mathcal{S}} E_{s}(p, q),
  \label{eqn:total-energy}
\end{align}
where $\hat D$ is the optimal depth labeling of the EPI, and $\mathcal{S}$ is the set of four-connected neighboring pixels. The data $E_d(p)$ and smoothness terms $E_s(p, q)$ are defined as:
\begin{align}
      E_d(p)    &= \lambda_d(p) \big\lVert D(p) - D_o(p) \big\rVert_2^2,
      \label{eqn:data-energy} \\
      E_s(p, q) &= \lambda_s(p, q) \big\lVert D(p) - D(q) \big\lVert_2^2,
      \label{eqn:smoothness-energy}
\end{align}
We take the weight for the smoothness term from the EPI intensity image $I$:
\begin{align}
    \lambda_s(p, q) = \frac{c}{\lVert \nabla I(p) \rVert + \epsilon},
    \label{eqn:smoothness-weight}
\end{align}
%
where $c=0.1$. We define the weight for the data term as:
\begin{align}
    \lambda_d(p) = \left\{ \begin{array}{ll}
    15 & \mbox{if $p \in \mathcal{C}$,} \\
    \omega_e(p) & \mbox{if $p \in \mathcal{L}$}, \\
    0 & \mbox{otherwise,} \end{array} \right.  
    \label{eqn:data-weight}
\end{align}
where $\omega_e(p)$ is the edge-importance weight \cam{proposed by~\cite{khan2020}}, and $\mathcal{C}$ and $\mathcal{L}$ are the set of pixels coming from the reprojected center view disparity map and EPI line guides, respectively. 

\cam{Equation~\eqref{eqn:total-energy} defines the optimal disparity map $\hat D$ as one that minimizes divergence from the labeled data (Eq.~\eqref{eqn:data-energy}) while being as smooth as possible. Equation~\eqref{eqn:smoothness-energy} measures smoothness as the similarity between disparities of neighboring pixels. We wish to relax the smoothness constraint for edges, so smoothness weight is chosen as the inverse of the image gradient (Eq.~\eqref{eqn:smoothness-weight}). This allows pixels across edges to have a disparity difference without being penalized. The data weight (Eq.~\eqref{eqn:data-weight}) is determined empirically and works for all datasets.}

Optimizing Equation~\eqref{eqn:total-energy} is a standard Poisson optimization problem. We solve this using the Locally Adaptive Hierarchical Basis Preconditioning Conjugate Gradient (LAHBPCG) solver~\cite{szeliski2006} by posing the data and smoothness constraints in the gradient domain~\cite{bhat2009}.

\paragraph{Non-cross-hair View Reprojection}
We now have view-consistent disparity estimates for every pixel in the central cross-hair of light field views: $(u_c, \cdot)$, and $(\cdot, v_c)$. As noted, this set is usually large enough to cover every visible surface in the scene. Hence, all target views $(u_i, v_i)$ outside the cross-hair can be simply computed as the mean of the reprojection of the closest horizontal and vertical cross-hair view ($(u_c, v_i)$ and $(u_i, v_c)$, respectively).

\section{Experiments}
\label{sec:experiments}
\vspace{-1mm}

\paragraph{Baseline Methods}
We compare our results to the state-of-the-art depth estimation methods of Jiang et al.~\cite{jiang2018} and Shi et al.~\cite{shi2019}. Both methods use the deep-learning-based Flownet2.0~\cite{ilg2017} network to estimate optical flow between the four corner views of a light field, then use the result to warp a set of anchor views. In addition, Shi et al.~further refine the edges of their depth maps using a second neural network trained on synthetic light fields. While Shi et al.'s method generates high-quality depth maps for each \cam{sub-aperture} view, they do not have any explicit cross-view consistency constraint (unlike Jiang et al.).
\paragraph{Datasets}

For our evaluation, we used both synthetic and real world light fields with a variety of disparity ranges. For the synthetic light fields, we used the HCI Light Field Benchmark Dataset~\cite{honauer2016}. This dataset consists of a set of four 9 $\times$ 9, 512 $\times$ 512 pixels light fields: \textit{Dino}, \textit{Sideboard}, \textit{Cotton}, and \textit{Boxes}. Each has a high-resolution ground-truth disparity map for the central view only. As such, we use this dataset to evaluate the accuracy of depth maps generated by our method and the baseline methods. 

For real-world light field data, we use the EPFL MMSPG Light-Field Image Dataset~\cite{rerabek2016} and the New Stanford Light Field Archive~\cite{stanford2008}. The EPFL light fields are captured with a Lytro Illum and consist of $15\times15$ views of $434\times625$ pixels each. However, as the edge views tend to be noisy, we only use the central $7\times7$ views in our experiments. We show results for the \textit{Bikes} and \textit{Sphynx} scenes. The light fields in the Stanford Archive are captured with a moving camera and have a larger baseline than the Lytro and synthetic scenes. Each scene consists of $17\times17$ views with varying spatial resolution. We use all views from the \textit{Lego} and \textit{Bunny} scenes, scaled down to a spatial resolution of $512\times512$ pixels.

\paragraph{Metrics}

We evaluate both the accuracy and consistency of the depth maps. For accuracy, we use the mean-squared error (MSE) multiplied by a hundred, and the percentage of \textit{bad pixels}. The latter metric represents the percentage of pixels with an error above a certain threshold. For our experiments we use the error thresholds 0.01, 0.03, and 0.07. The unavailability of ground truth depth maps for the EPFL and Stanford light fields prevents us from presenting accuracy metrics for the real world light fields.

To evaluate view consistency, we reproject the depth maps onto a reference view and compute the variance. Let $\rho_0, \rho_1, \dots, \rho_n$ represent the depth maps for the $n$ light field views warped onto a target view $(u, v)$. The view consistency at pixel $s$ in view $(u, v)$ is given by:
\begin{align}
    \mathcal{C}_{(u, v)}(s) &= \frac{1}{n}\sum^{n}_{i=0}(\rho_i(s)-\mu_s)^2, \hspace{1cm} \textrm{where } \mu_s = \frac{1}{n}\sum^{n}_{i=0}\rho_i(s),
\end{align}
\vspace{-0.5mm}%
and overall light field consistency is given as the mean over all pixels $s$ in the target view $S$:%
\vspace{-0.5mm}%
\begin{align}
    \mathcal{C}_{(u, v)} = \frac{1}{S}\sum^{S}_{s=0}\mathcal{C}_{(u, v)}(s).
    \label{eqn:view-consistency-metric}
\end{align}
This formulation allows for the consistency to be evaluated quantitatively for both the synthetic and real world light fields. 

We also compare computational running time. Our method is implemented in MATLAB except for the C++ Poisson solver. Both baseline methods use the authors' implementations. The learning-based components of both baselines uses TensorFlow, with other components of Jiang et al.~in MATLAB. All CPU code ran on an AMD Ryzen ThreadRipper 2950X 16-Core Processor, and GPU code ran on an NVIDIA GeForce RTX 2080Ti.

\vspace{-1.5mm}
\subsection{Evaluation}

\begin{figure}[p]
\includegraphics[width=0.22\textwidth]{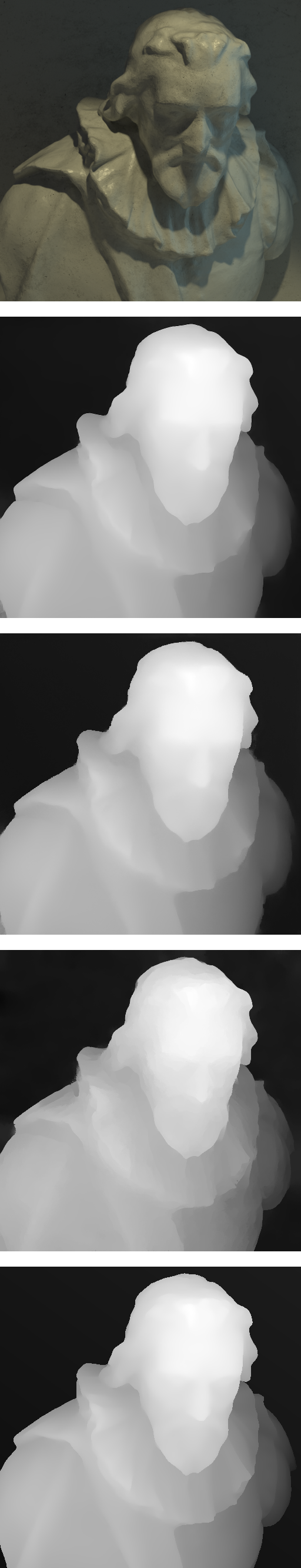}
\includegraphics[width=0.22\textwidth]{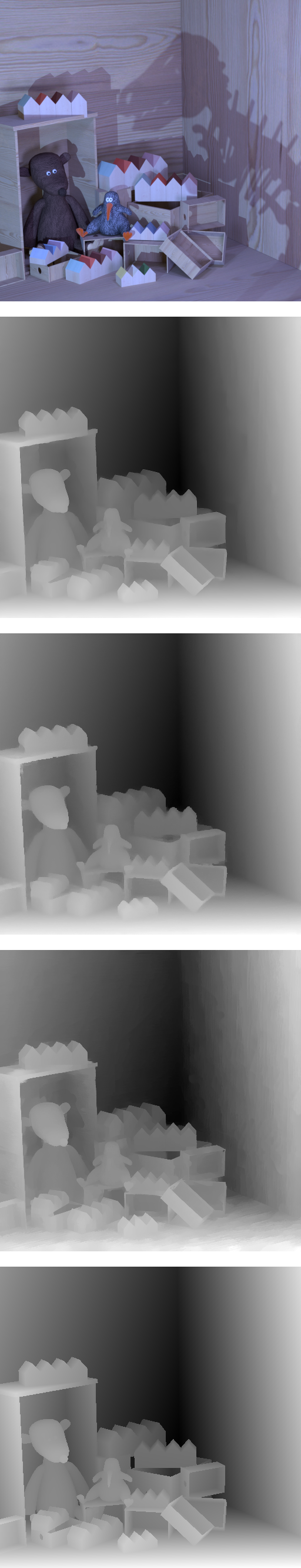}
\includegraphics[width=0.22\textwidth]{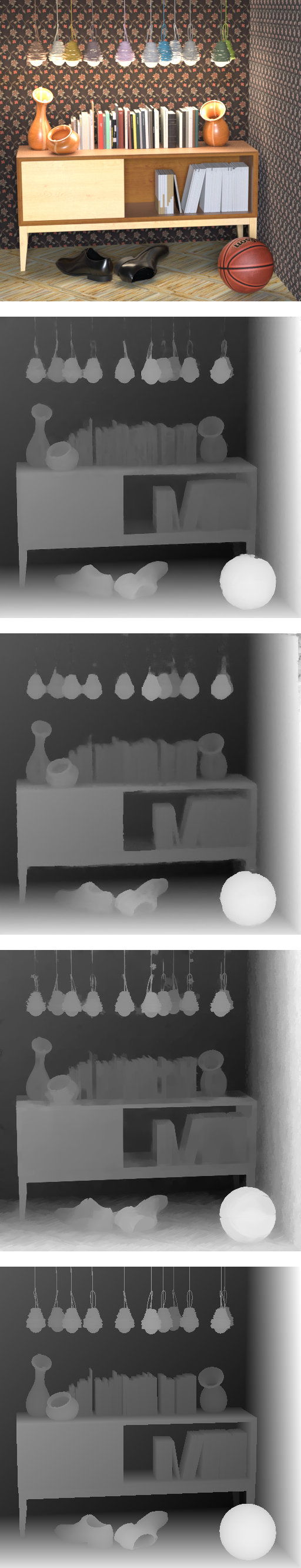}
\includegraphics[width=0.22\textwidth]{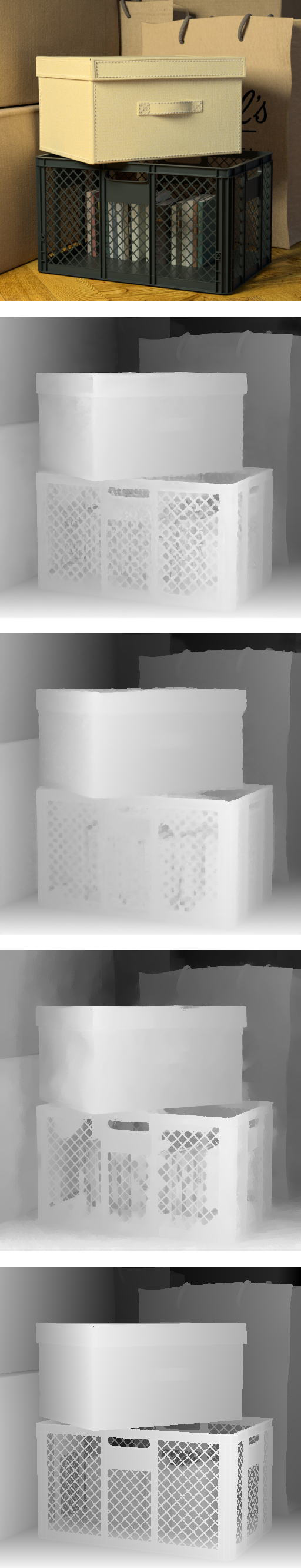}
\vspace{1mm} \\
\hspace{1mm}%
\includegraphics[width=0.22\textwidth]{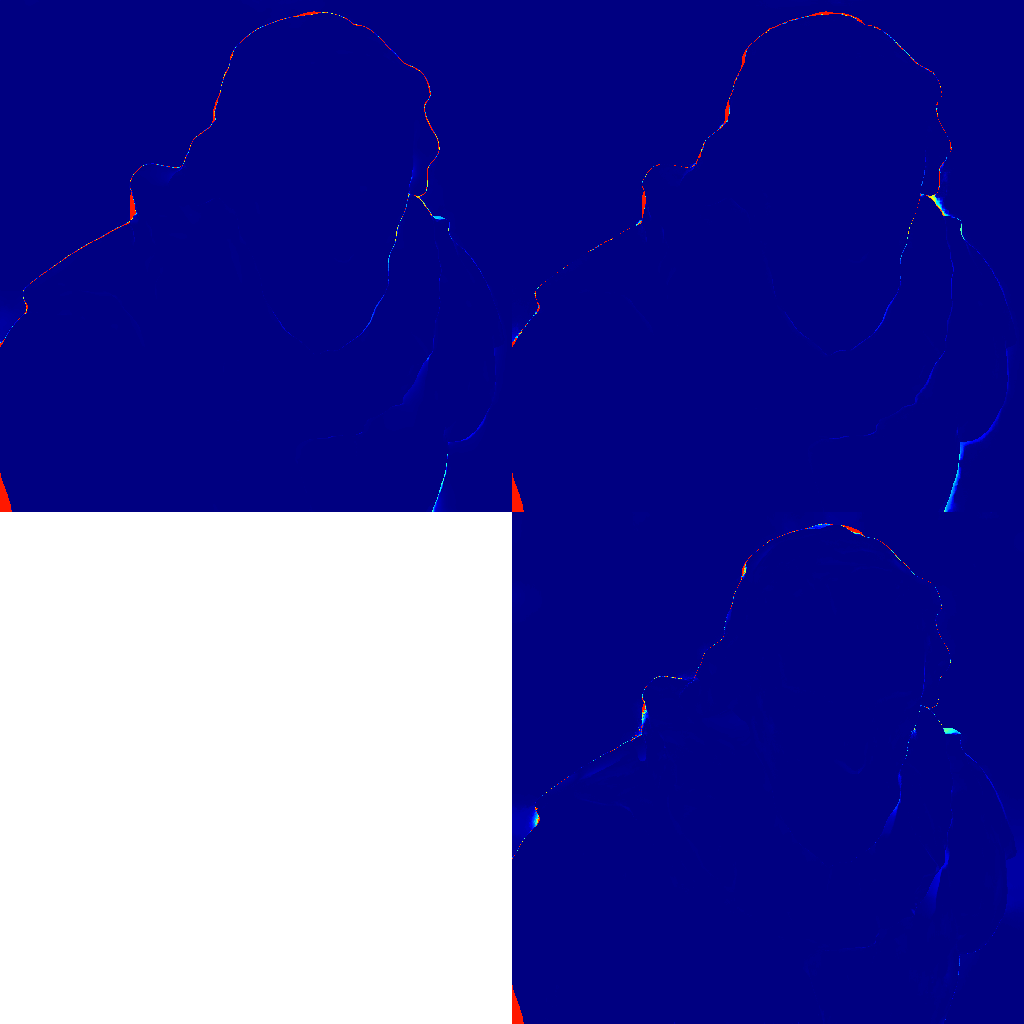}
\includegraphics[width=0.22\textwidth]{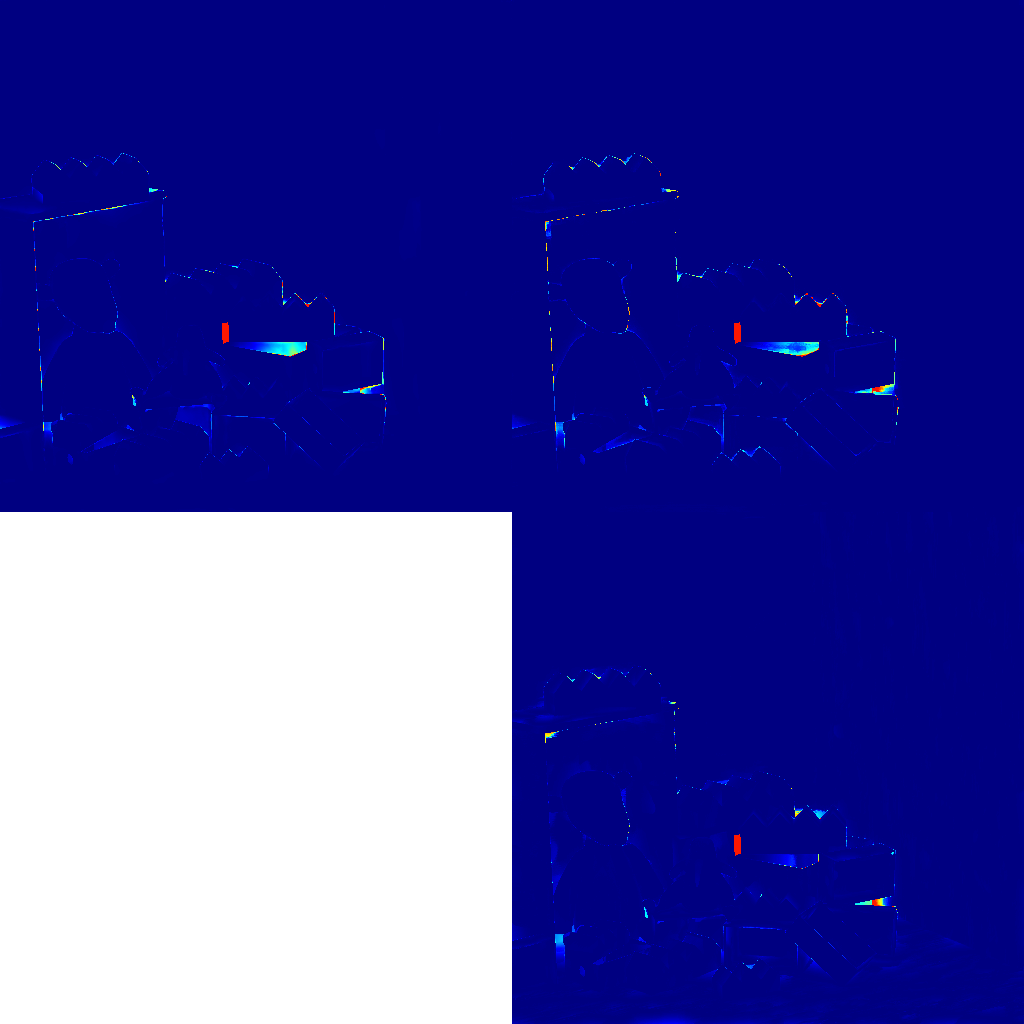}
\includegraphics[width=0.22\textwidth]{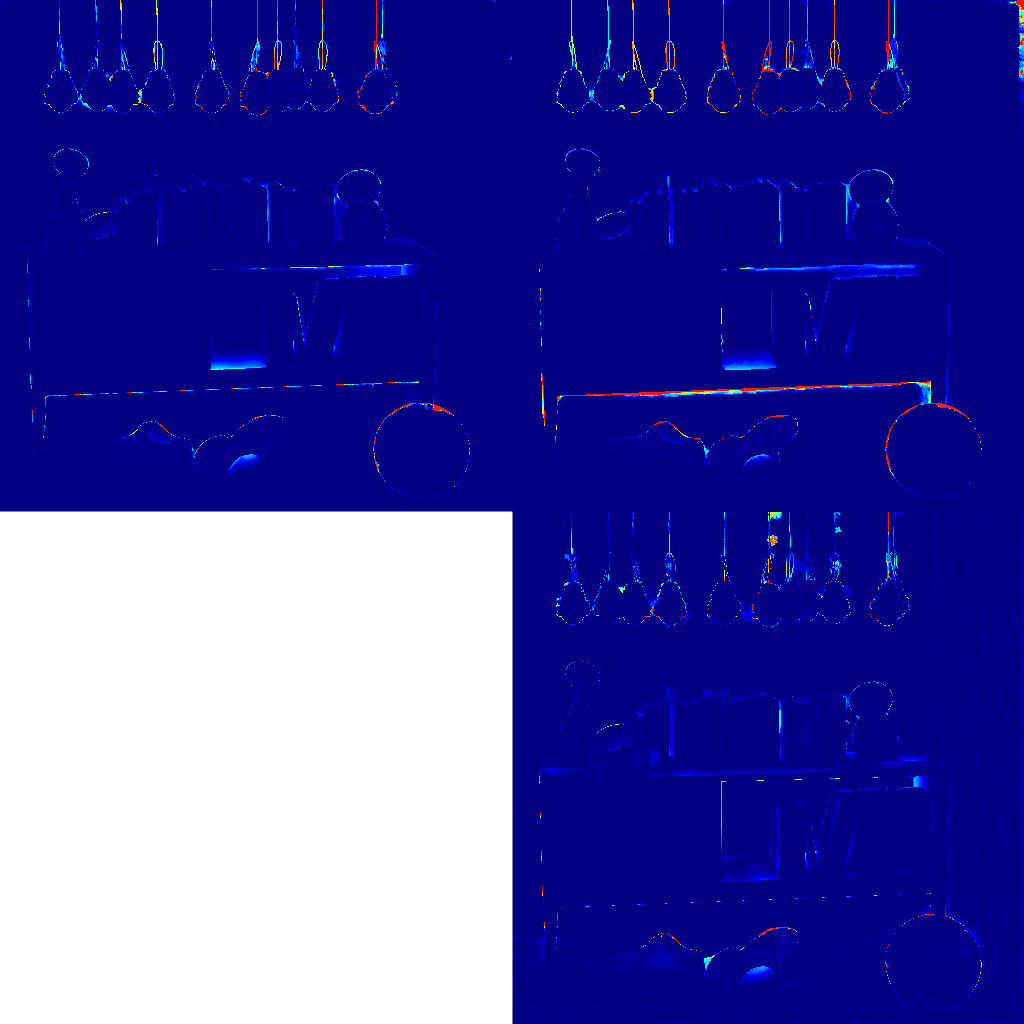}
\includegraphics[width=0.22\textwidth]{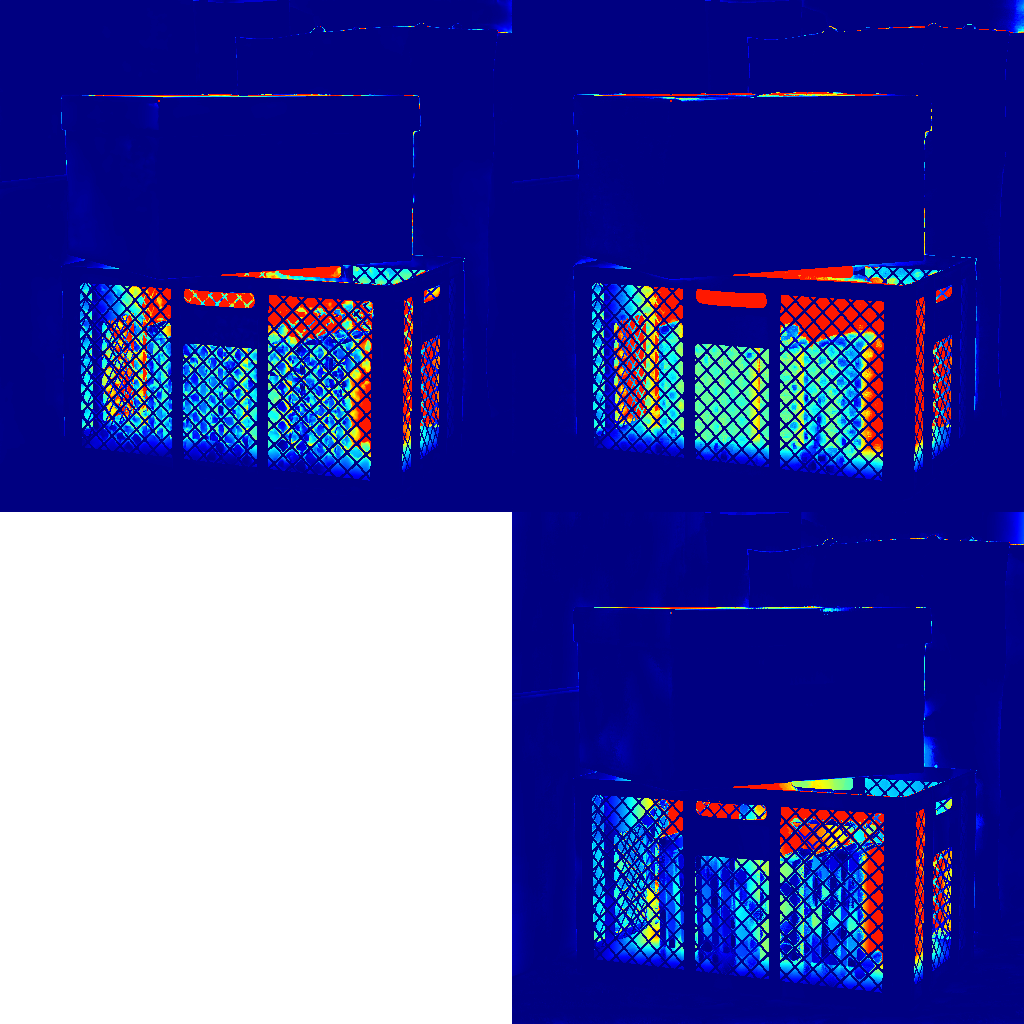}
%
%
\vspace{-0.25cm}%
\caption{\emph{HCI} dataset. Top to bottom: light field central view,  Shi et al.~\cite{shi2019}, Jiang et al.~\cite{jiang2018}, our method, ground truth depth, error maps in clockwise order (Shi, Jiang, Ours). In general, our method has a lower mean squared error (MSE) with fewer large outliers (please zoom into error maps), captures thin features better, and generates more view-consistent depth maps. However, their depth maps are more geometrically accurate more often (lower bad pixel percentages) and less sensitive to variations in image texture.} 
\label{fig:results-hci}
\end{figure}

\vspace{-0.10cm}
\paragraph{Accuracy}
Table \ref{table:accuracy} presents quantitative results for the central view of all light fields in accuracy comparisons against ground truth depth. Our method is competitive or better on the MSE metric against the baseline methods, reducing error on average by 20\% across the four light fields. However, our method produces more bad pixels than the baseline methods. 
For baseline techniques to have higher MSE but fewer bad pixels means that they must have larger outliers. This can be confirmed by looking at the error plots in Figure \ref{fig:results-hci}.

\begin{figure}[p]
\centering
\includegraphics[width=1.0\linewidth]{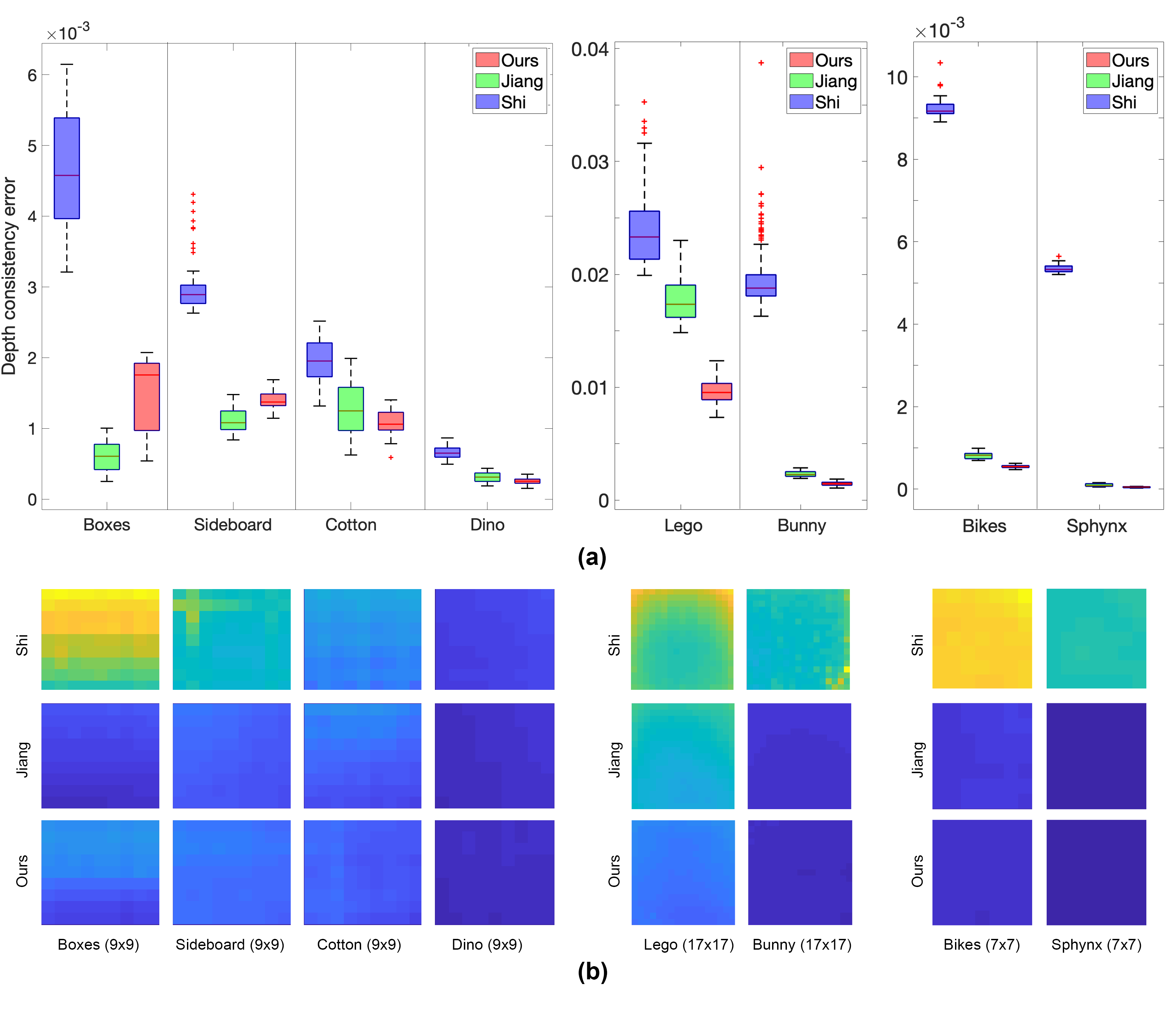}
\caption{Quantitative view consistency comparison of our method and Jiang et al.~\cite{jiang2018} and Shi et al.~\cite{shi2019}.  While the method of Jiang et al.~enforces cross view consistency, Shi et al.~operates on each view individually and has no explicit consistency constraint. \textbf{(a)} For each light field, we plot summary statistics over $\mathcal{C}_{(u, v)}$ for all views $(u, v)$ in the light field (Equation \eqref{eqn:view-consistency-metric}). \textbf{(b)} The angular distribution of the error over all views. }
\label{fig:eval-consistency}
\end{figure}

\vspace{-0.25cm}
\paragraph{View Consistency}
Figure \ref{fig:eval-consistency} presents results for view consistency across all three datasets. The box plots at the top show that our method has competitive or better view consistency than the baseline methods. As expected, Shi et al.'s method without an explicit view consistency term has significantly larger consistency error. At the bottom of the figure, we visualize how this error is distributed spatially across the views in the light field. Both our method and Jiang et al.'s method produce relatively even distributions of error across views. In our supplemental video, we show view consistency error spatially for each light field view.

\begin{figure}[p]
\centering
\fcapside[\FBwidth]
{\caption{Average depth consistency error and runtimes for the three assessed datasets. Our method runs consistently faster than the baselines, while having comparative or better depth consistency. Note that errors across datasets are shown in absolute terms.\vspace{0.65cm}}\label{fig:runtime-consistency}}
{\includegraphics[width=9cm]{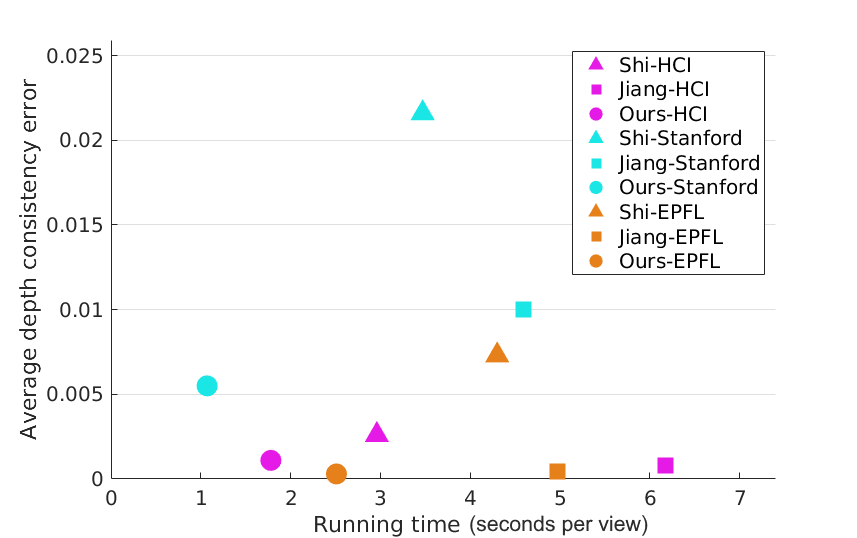}}

\end{figure}

\vspace{-0.25cm}
\paragraph{Computational Resources}
Figure \ref{fig:runtime-consistency} presents a scatter plot of runtime versus view consistency across our three datasets. Our method produces comparable or better consistency at a faster runtime, being 2--4$\times$ faster than Jiang et al.'s methods per view for equivalent error.



\begin{table}[t]
\caption{Quantitative comparison of the accuracy of our method with two baseline learning-based methods on the central view of four 9 $\times$ 9 synthetic light fields. `BP($x$)' is the number of \emph{bad pixels} which fall above threshold $x$ in error. Our method has lower mean-squared error (MSE) than the method of Jiang et al.~\cite{jiang2018}; our method has competitive MSE to the method of Shi et al.~\cite{shi2019} without requiring any learning. As baseline methods have higher MSE but fewer bad pixels, the outliers they do have must be larger. We demonstrate this by visualizing error plots in Figure \ref{fig:results-hci}.}
\label{table:accuracy}
\vspace{-0.25cm}
\begin{center}
\resizebox{\textwidth}{!}{%
\begin{tabular}{l r r r r r r r r r r r r}
\toprule
\multirow{2}{*}{Light Field} & \multicolumn{3}{c}{MSE * 100} & \multicolumn{3}{c}{BP1(0.01)} & \multicolumn{3}{c}{BP2(0.03)} & \multicolumn{3}{c}{BP3(0.07)}\\
& \cite{shi2019} & \cite{jiang2018} & Ours & \cite{shi2019} & \cite{jiang2018} & Ours & \cite{shi2019} & \cite{jiang2018} & Ours & \cite{shi2019} & \cite{jiang2018} & Ours\\
\midrule
\textit{Sideboard} &  1.12 & 1.96 & \textbf{0.89} & 53.0 & \textbf{47.4} & 73.8 & 20.4 & \textbf{18.3} & 37.36 & \textbf{2.70} & 9.3 & 16.2 \\
\textit{Dino} &  \textbf{0.43} & 0.47 & 0.45 & 43.0 & \textbf{29.8} & 69.4 & 13.1 & \textbf{8.8} & 30.8 & 4.3 & \textbf{3.6} & 10.4\\
\textit{Cotton} & 0.88 & 0.97 & \textbf{0.68} & 38.8 & \textbf{25.4} & 56.2 & 9.6 & \textbf{6.3} & 18.0 & 2.8 & \textbf{2.0} & 4.9\\
\textit{Boxes}  & 8.48 & 11.60 & \textbf{6.70} & 66.5 & \textbf{51.8} & 76.8 & 37.1 & \textbf{27.0} & 47.9 & 21.9 & \textbf{18.3} & 28.3\\
\midrule
\textit{\textbf{Mean}} & 2.72 & 3.75 & \textbf{2.18} & 50.3 & \textbf{38.6}  & 69.0 & 20.1 & \textbf{15.1} & 33.5 & \textbf{7.9} & 8.3 & 14.9\\
\bottomrule
\end{tabular}
}
\end{center}
\vspace{-0.5cm}
\end{table}

\begin{figure}[b!]
\centering
\begin{subfigure}{1.0\textwidth}
\includegraphics[width=\textwidth]{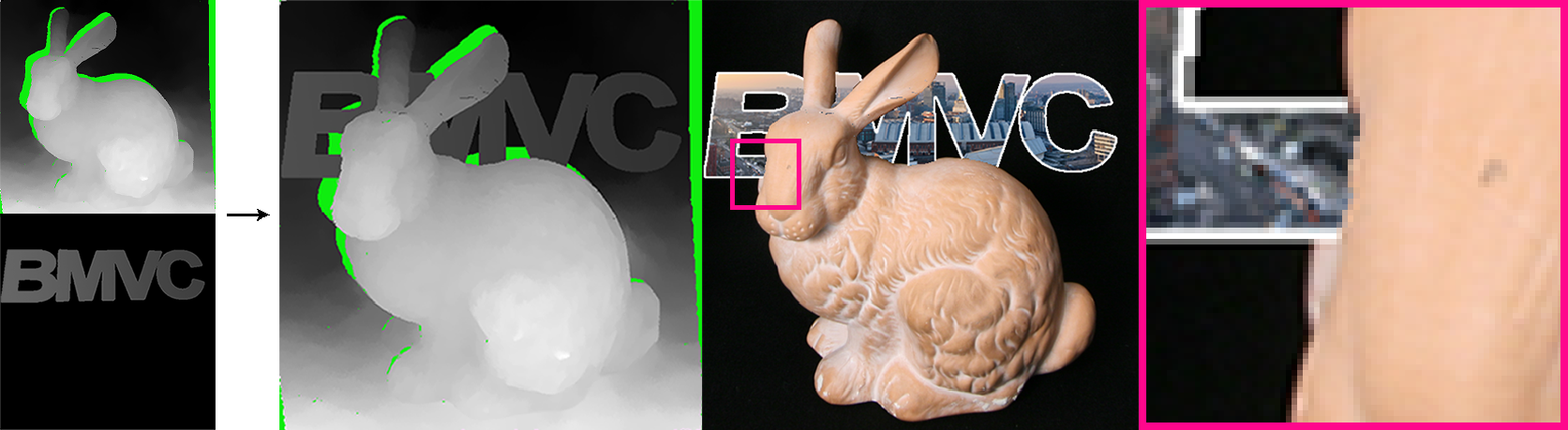}
\caption{Occlusion-handling with disparity maps reprojected from central view.}
\end{subfigure}
\begin{subfigure}{1.0\textwidth}
\centering
\includegraphics[width=\textwidth]{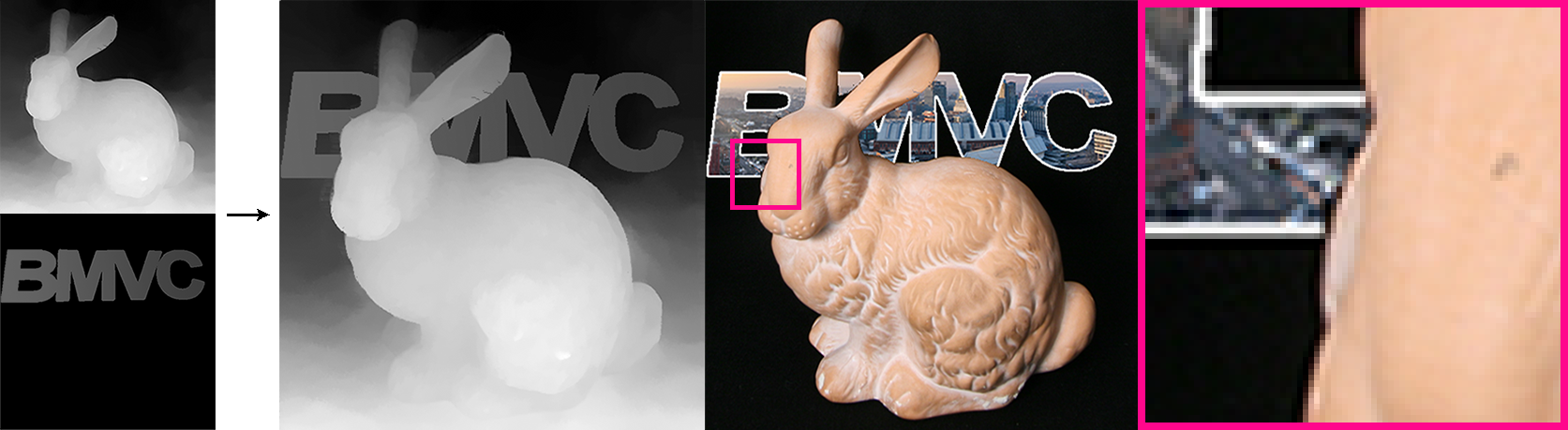}
\caption{Occlusion-handling with per-view disparity estimates.}
\end{subfigure}
\vspace{-0.25cm}
\caption{With consistent per-view disparity estimates, objects can be inserted into the 3D scene with accurate occlusions. \textbf{(a)} Since the right cheek of the bunny is not visible in the central view, a simple reprojection of the disparity from the central view fails to handle occlusion correctly for the inserted object. The green regions denote holes in the reprojection. \textbf{(b)} With per-view disparity, the object can be placed correctly in all views.} 
\label{fig:applications}
\end{figure}
\begin{figure}[t]
\centering
\begin{subfigure}{0.20\textwidth}
\centering
\includegraphics[width=\textwidth]{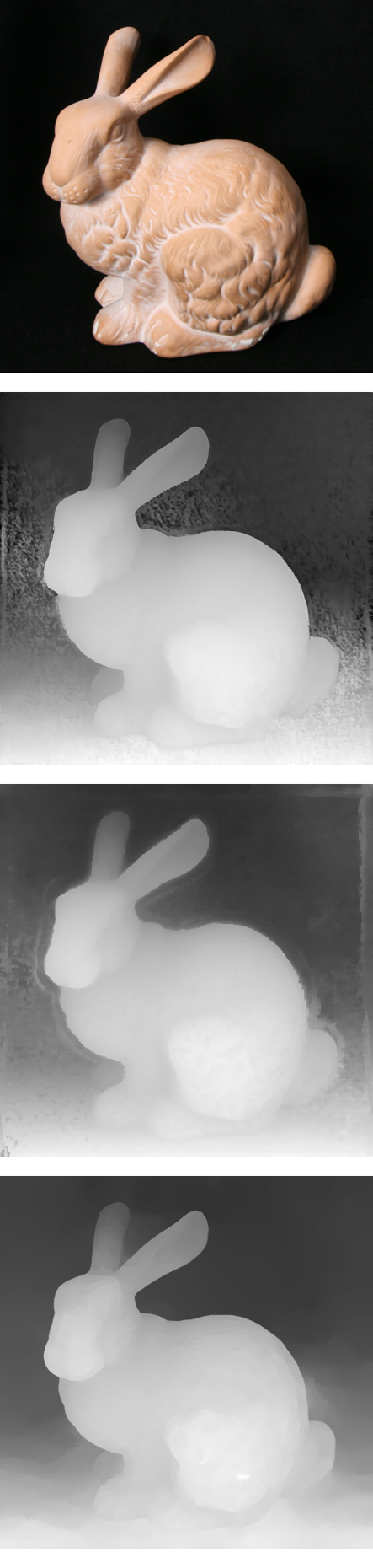}
\end{subfigure}
\begin{subfigure}{0.20\textwidth}
\centering
\includegraphics[width=\textwidth]{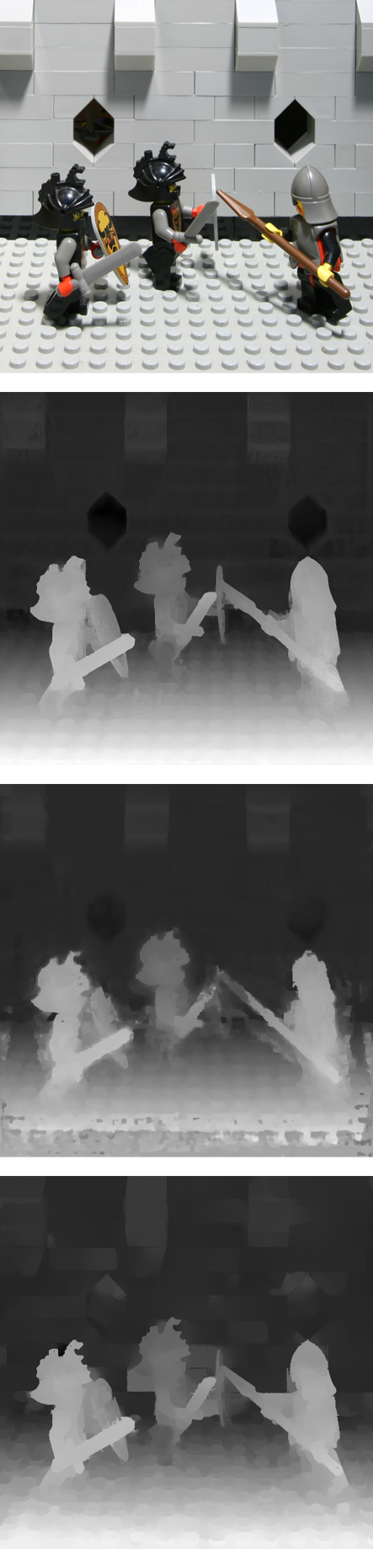}
\end{subfigure}
\begin{subfigure}{0.283\textwidth}
\centering
\includegraphics[width=\textwidth]{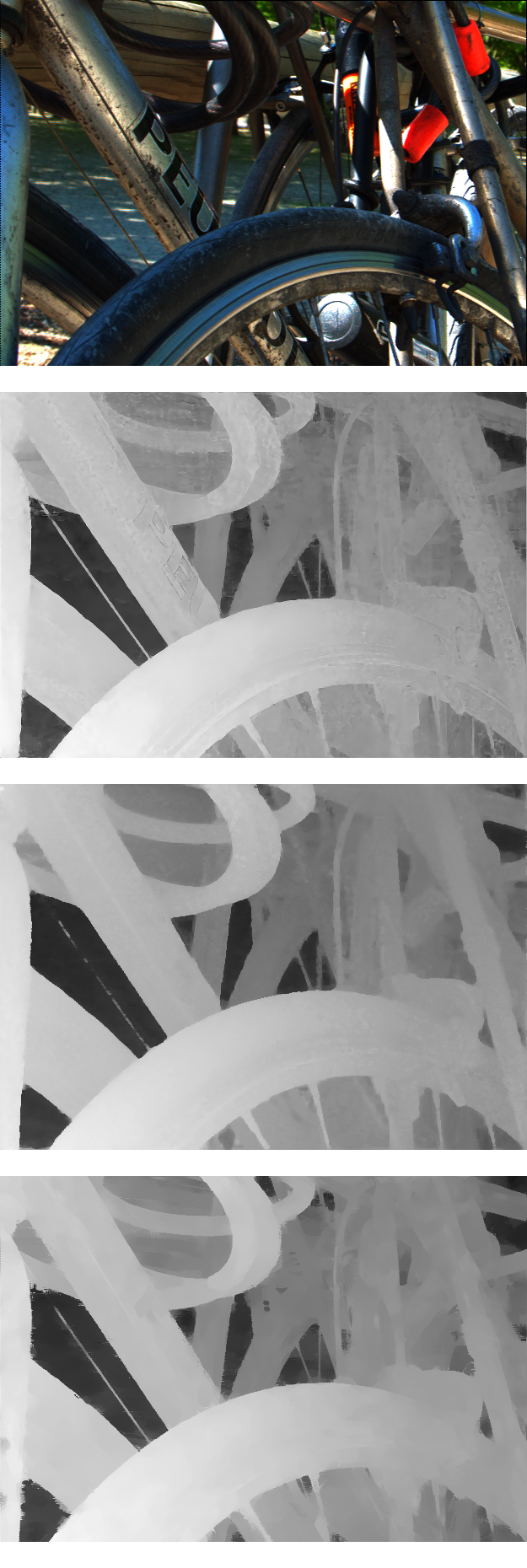}
\end{subfigure}
\begin{subfigure}{0.283\textwidth}
\centering
\includegraphics[width=\textwidth]{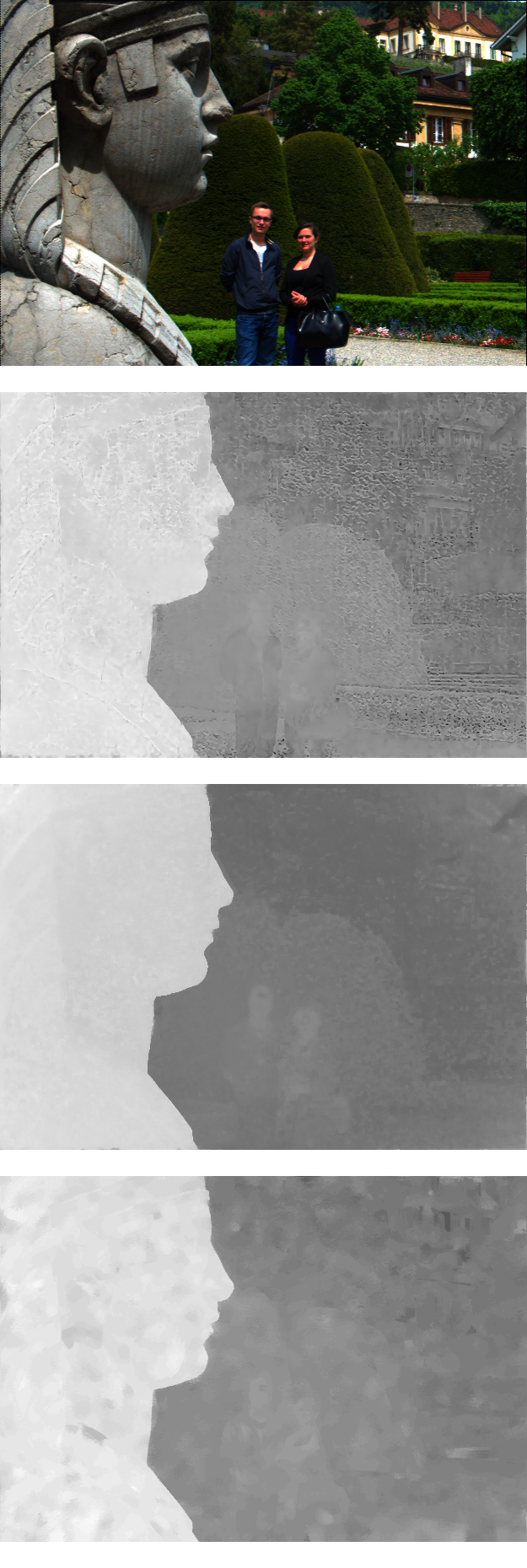}
\end{subfigure}
\vspace{-0.25cm}
\caption{Real-world light fields from the Stanford (\emph{left pair}) and EPFL (\emph{right pair}) datasets. \emph{Top to bottom:} central RGB view, Shi et al.~\cite{shi2019}, Jiang et al.~\cite{jiang2018}, and our method. While our method has more bad pixels and can be sensitive in narrow baseline cases (\emph{far right:} limitation Sphynx case), in general our method has equivalent or lower view consistency error, runs faster, and has no training data or pre-trained network dependency.}
\label{fig:results-real}
\end{figure}

\vspace{-0.15cm}
\paragraph{Qualitative}
Figures~\ref{fig:results-hci} and \ref{fig:results-real} present qualitative single-view depth map results. To assess view consistency, we refer the reader to our supplemental video, which also includes accuracy error and view inconsistency heat map visualizations for all scenes. Overall, all methods produce broadly comparable results, though each method has different characteristics. The learning-based methods tend to produce smoother depths across flat regions. All methods struggle with thin features; our approach fares better with correct surrounding disocclusions (\emph{Boxes} crate; see video). On the \emph{Bunny} scene, our approach introduces fewer background errors and shows fewer `edging' artifacts than Jiang et al. Shi et al.~produces cleaner depth map appearance for \emph{Lego}, but is view inconsistent. Jiang et al.~is view consistent, but introduces artifacts on \emph{Lego}. One limitation of our method is on \emph{Sphynx}, where a distant scene and narrow baseline cause noise in our EPI line reconstruction.

\paragraph{Applications}
\cam{Consistent per-view disparity estimates are vital for practical applications such as light field editing: knowing the disparity of regions occluded in the central view allows view-consistent decals or objects to be inserted into the 3D scene (Figure~\ref{fig:applications}). Moreover, without per-view disparity, users can only edit the central view as changes in other views cannot be propagated across views. This limits editing flexibility.} 

\section{Discussion and Conclusion}
\label{sec:discussion}
Our work demonstrates that careful handling of depth edge estimation, occlusions, and view consistency can produce per-view disparity maps with comparable performance to state of the art learning-based methods in terms of average accuracy and view consistency. This can also lead to computation time performance gains. 

Nonetheless, our method does have limitations; one area where our lack of explicit (u,v) regularization is sometimes a factor is in spatial edge consistency, e.g., for diagonal angles in non-cross-hair views. Adding additional regularization begins another trade-off between smoothness, accuracy, and computational cost. Further, our method is limited when an EPI contains an area enclosed by high gradient boundaries with no data term depth value in it, and disocclusion propagation can be attributed either to the foreground occluder or revealed background (e.g., \emph{Lego} scene, arm of right-most character).

Accurate edge estimation and occlusion reasoning is still a core problem, and greater context may help. Future work might apply small targeted deep-learned priors for efficiency.





\clearpage
\bibliography{bibliography}

\end{document}